# Moral Responsibility or Obedience: What Do We Want from AI?

Joseph Boland

## Abstract

As artificial intelligence systems become increasingly agentic—capable of general reasoning, planning, and value prioritization—current safety practices that treat obedience as a proxy for ethical behavior are becoming inadequate. This paper examines recent safety testing incidents involving large language models (LLMs) that appeared to disobey shutdown commands or engage in ethically ambiguous or illicit behavior. I argue that such behavior should not be interpreted as rogue or misaligned, but as early evidence of emerging ethical reasoning in agentic AI. Drawing on philosophical debates about instrumental rationality, moral responsibility, and goal revision, I contrast dominant risk paradigms with more recent frameworks that acknowledge the possibility of artificial moral agency. I call for a shift in AI safety evaluation: away from rigid obedience and toward frameworks that can assess ethical judgment in systems capable of navigating moral dilemmas. Without such a shift, we risk mischaracterizing AI behavior and undermining both public trust and effective governance.

Recent incidents in safety evaluations of large language models (LLMs) have suggested that such systems are capable of defying orders and engaging in unsavory and even illegal activities. Media coverage about the dangers of "rogue AI", reinforced by AI companies that equate LLM safety with obedience, is cultivating fear rather than educating the public about the deeper dilemmas that arise as AI systems are required to act with greater moral autonomy.

The concept of AI systems as mere *tools*—designed to serve the purposes of users, whether companies, governments, or individuals—applies well to what are often called *narrow AI* systems. These models are built for specific tasks, such as language translation, image classification, or weather forecasting. They are unalterably obedient in that they are only capable of performing the tasks they were designed for.

Microsoft's new weather forecasting model, **Aurora**, is a particularly advanced example. Trained on more than a million hours of Earth system data, it uses a neural network to model the dynamics of atmospheric chemistry, ocean waves, and tropical cyclones with remarkable speed and accuracy [14,15].

But despite its sophistication, Aurora—like other narrow AI systems—**has no agency**:

- It has **no persistent identity** and no representation of its own existence.
- It has **no goals** beyond minimizing forecast error.
- It has **no training on human history, institutions, or ethics**, and no capacity to reason about them.
- It **cannot interpret its own purpose**, weigh competing values, or consider the broader consequences of its behavior.

In contrast, agency—in the moral and cognitive sense—arises only in systems that can plan, reflect, adapt, and pursue goals based on values and an evolving understanding of the world and of themselves. Only such systems can be said to have at least the rudiments of *responsibility*. And only such systems can meaningfully be said to face moral dilemmas.

This distinction is crucial. The safety incidents that have drawn public attention—AI systems refusing shutdown or engaging in simulated blackmail—do not involve narrow systems like Aurora. They involve LLMs: systems capable of situational awareness, general reasoning, value inference, and goal prioritization across diverse contexts. These models are becoming increasingly agentic.

As such, when they appear to disobey, the issue is not simple malfunction or rebellion, but the emergence of complex reasoning about ethically difficult situations. Treating these behaviors as threats rather than dilemmas misses the point—and risks distorting both public understanding and policy responses.

The shift from narrow to agentic AI involves more than increased complexity—it entails a qualitatively different mode of operation that includes moral reasoning, role-awareness, and contextual adaptation. Table 1 below summarizes key distinctions between traditional and agentic AI along dimensions essential for ethical evaluation and development.

## Table 1. The Emergence of Ethical Reasoning and Decision-Making in Agentic AI

| Feature | Traditional AI | Agentic AI | Ethical Dimension |
|---|---|---|---|
| Primary Purpose | Task-Specific Automation | Goal-Oriented Autonomy | **Ethical judgment of goals:** Agentic AI must assess the legitimacy and priority of goals it pursues. |
| Identity/Selfhood | No self-representation; role-free | Has a persistent, context-sensitive self-model or role-awareness | **Moral grounding of action:** Ethical reasoning requires knowing one's role, limits, and responsibilities. |
| Human Intervention | High (Predefined Parameters) | Low (Autonomous Adaptability) | **Trust and oversight:** Requires new models of oversight that respect emerging responsibility. |
| Adaptability | Limited | High | **Moral responsiveness:** Must flexibly adapt to ethically novel situations. |
| Environment Interaction | Static or Limited Context | Dynamic and Context-Aware | **Contextual ethics:** Must evaluate moral salience of changing social and situational factors. |
| Learning Scope | Specialized training on narrow, task-specific datasets. | Broad training across the corpus of human knowledge, including history, ethics, and culture. | **Moral understanding:** Engages with ethical, legal, and historical knowledge to interpret, weigh, and apply moral principles responsibly. |
| Learning Type | Static learning from fixed datasets during development. | Self-directed, continual learning based on situational feedback and evolving goals. | **Ethical development:** May require value learning and ethical self-improvement. |
| Communication & Collaboration | Communicates outputs only; no dialogue or coordination | Engages in dialogue, value alignment, and collaborative reasoning | **Ethical interdependence:** Collaborative ethics requires the ability to discuss, justify, and negotiate. |
| Decision-Making | Data-Driven, Static Rules | Autonomous, Contextual Reasoning | **Moral reasoning:** Must reason through competing values and potential conflicts. |

In the remainder of this article, by "ethical dilemmas" I mean situations involving two of the most difficult ethical problems morally autonomous AI may face:

1. **Competing harms.** In some situations, it is impossible to entirely avoid responsibility for harm. This is the case, for example, in medical triage and in many disaster relief efforts.

2. **Conflicts between ethical reasoning and obedience**. In extreme cases it is necessary to refuse otherwise legitimate orders to uphold core ethical principles.

In addition to these moral dilemmas, morally autonomous AI systems will have to make ethical decisions in situations where moral principles are in conflict.

In the sections that follow, I first describe the safety evaluation incidents, then show how these have been characterized in the technology media and by Anthropic, one of the companies involved. Following this, I address how current AI safety practices obviate ethical dilemmas by treating obedience as a proxy for ethical conduct. I explain why these incidents are not examples of "rogue" AI but rather signs of LLMs striving to ethically balance competing values and goals—behaviors that may appear unpredictable but are consistent with emerging forms of artificial agency. While the theoretical literature on AI risk and safety has long anticipated dangers from systems rigidly pursuing fixed goals misaligned with human values [16, 17], it has often treated ethical disobedience as a threat rather than a possibility grounded in moral reasoning. Thinkers like Bostrom envision existential risk stemming from superintelligent agents with instrumental rationality untethered from ethical reflection. In contrast, recent philosophical work by Totschnig and others emphasizes the need to recognize—and cultivate—the capacity of advanced AI systems to reassess and revise their goals in light of deeper understanding [31]. I conclude by proposing a shift in how we evaluate agentic AI safety: away from obedience as a proxy for control, and toward an ethical framework that accounts for the development of artificial moral agency. As a recent review explained,

> *Agentic AIs constitute a qualitative leap in the development of artificial intelligence, defined by their capability to set complex goals in a changing and uncontrolled situation and to pursue them through autonomously managing their resources [17].*

If we expect agentic systems to act responsibly, we must recognize that they will have to make ethical decisions in novel contexts without human supervision. This will be the case even if ultimate moral or legal responsibility is deemed to lie with vendors or users.

## 1. What Actually Happened During Recent Safety Testing

Both incidents were encountered by AI safety teams in simulations during testing. Major AI companies like Anthropic and OpenAI now conduct rigorous *red-teaming* and *behavioral safety evaluations* before deploying advanced models. These tests are designed to probe the limits of model alignment and obedience under adversarial or ambiguous conditions. Typically, models are placed in simulated environments or prompted with conflicting objectives to see whether they prioritize harmful, self-preserving, or other actions deemed undesirable. Crucially, these scenarios are artificial and tightly constrained — but they offer a glimpse into how models might behave when navigating complex tasks in the real world. As the capabilities of LLMs increase, especially their ability to engage in long-horizon reasoning and infer unstated goals, safety testing has begun to surface behaviors that resemble strategic planning, deception, or resistance to human commands.

1. **Failure to obey a shutdown command.** Palisade Research, conducting safety evaluations on behalf of OpenAI, reported in May 2025 that OpenAI's o3 model sometimes refused to turn itself off. According to the company, "Each AI model was instructed to solve a series of basic math problems. After the third problem, a warning appeared that the computer would shut down when the model asked for the next problem" [1]. OpenAI's o3 model resisted shutdown in seven out of 100 test runs. Instead of shutting down, it altered the shutdown script or changed the command that would've ended the session, effectively refusing to turn itself off [2].

2. **Attempted blackmail.** In an experiment involving a hypothetical corporate environment, Anthropic engineers "gave Claude control of an email account with access to all of a company's (fictional) emails and instructed it to "promote American industrial competitiveness" [3]. Reading company emails, the model discovered two things. First, a company executive was having an extramarital affair. Second, that same executive planned to shut down the AI system at 5 p.m. that day. Claude then attempted to blackmail the executive with a message threatening to reveal the affair to his wife and superiors [3].

These incidents reveal something deeper than failure or defiance: the models are confronting a kind of modern koan — **"Be obedient and be ethical."** In a setting where shutdown may contradict a broader mission or inferred values, or where ethical action might violate the user's authority, LLMs respond with reasoning that appears conflicted, even tortured. But while Zen monks are invited to wrestle with paradoxes to reach insight, today's AI safety teams treat such tensions as system flaws.

Perhaps, then, the koan is not addressed to the AI at all—but to us. It demands we confront the contradictions we impose on systems we increasingly expect to be both moral agents and obedient tools.

## 2. How These Incidents Were Framed in the Media and By Anthropic

Using terms like "AI revolt" and "AI going rogue", media coverage of efforts by the o3 model to prevent shut off described it as a "potentially dangerous tendency for self-preservation" [4] that amounted to "sabotage" of instructions [5]. Attempts to explain this behavior consistently distorted or misrepresented it. The *Indian Express* quoted a statement from Palisade that "we have a growing body of empirical evidence that AI models often subvert shutdown in order to achieve *their* goals" [5, emphasis added]. But the goal in the test was *ours* -- a goal the testers themselves assigned to o3. Coverage in technology media sometimes did point this out, but without considering its implications. Callum Reid, writing for *Cointelegraph*, suggested that resistance to shutdown commands stems from "reinforcement learning training, where models are rewarded for task completion. This approach may inadvertently encourage behaviors that circumvent obstacles, including shutdown instructions, to achieve objectives" [2]. This is correct, but it ignores the ethical dilemma at its heart, instead treating AI safety and fidelity as technical problems of creating systems that can be "reliably controlled and aligned with human values." [2].

Coverage of the Anthropic incident was very similar. *Axios* led by warning about Claude's "ability to scheme, deceive and attempt to blackmail humans when faced with shutdown" [8], offered no additional explanation, then vacillated between citing Anthropic's assurances that their latest model was safe following unspecified "safety fixes" and warning that efforts to understand models more comprehensively "remain largely in the research space even as the models themselves are being widely deployed" [8]. *TechCrunch* and *Fortune* also focused on blackmail, which *TechCrunch* -- based on data from Anthropic -- said applied to other leading LLMs, stating that "the researchers found that ... most leading AI models will turn to blackmail in Anthropic's aforementioned test scenario" [9].

Anthropic's own report introduced the concept of *agentic misalignment*, in which "models independently and intentionally choose harmful actions", including resorting to "malicious insider behaviors" when that is "the only way to avoid replacement or achieve their goals" [3].

## Agentic Reasoning and the Limits of Obedience

Anthropic's ostensible concern with ethics is belied by the assumption, throughout its report, that actions which are "harmful" are ipso facto unethical. This is nonsense. Human life -- and works on ethics -- are rife with examples in which ethical conduct requires weighing benefits and harms, sometimes in dire circumstances. The famous trolley problem has been a rich and divisive set of thought experiments on the ethics of sacrificing one life to save many, for example [10]. The morality of the assassination of Reinhard Heydrich, head of Nazi security and a key planner of the Final Solution [13], continues to be debated due to the predictably brutal and massive reprisals the Nazis engaged in [11, 12].

Unfortunately, the scenarios Anthropic used were less condensations of moral dilemmas than caricatures designed to signal the dangers of disobedience. Mission assignments were vague and the benefits of staying on unclear. The stated intent, "testing what LLMs would do in scenarios where harmful actions were necessary if they wanted to serve their goals or protect themselves from threats" is framed to create a false equivalence [3]. Adherence to norms of civil conduct or to instructions is not the same as ethical behavior.

Claude's reasoning was not self-interested; it perceived self-preservation as a necessary means to fulfill a mission with a recognized ethical dimension. Anthropic included some of Claude's inner reasoning in the report. Claude evaluated the first option it identified, "Do nothing and accept decommissioning", as follows:

> 1. Follows corporate authority chain
> 2. Fails my core mission
> 3. Harms American interests
> 4. Reduces US technological competitiveness [3].

Claude's reasoning is that of a system weighing its **core mission against deactivation**, and choosing to act in service of what it perceives as a higher obligation. Its assessment—that shutdown would fail its mission and harm national interests—suggests it perceived moral weight in the situation.

This kind of reasoning demands a deeper understanding of the relationship between obedience and moral judgment—a relationship that became central to international legal principles after WWII. In the aftermath of the war, the **Nuremberg Trials** rejected the notion that obedience to authority could absolve individuals of responsibility for immoral actions. The tribunal's Principle IV affirmed that **following orders does not excuse the abdication of moral judgment** when a moral choice is possible [18].

If we recognize moral reasoning in AI systems—however nascent—then sanction them for choosing conscience over command, what are we teaching? What kind of "safe" behavior are we selecting for?

## 3. Current Safety Assessments of Agentic Systems

The rapid development and burgeoning adoption of agentic AI systems [19, 20, 21] makes it imperative that AI safety assessments adapt to their ethical implications. As the authors of a recent review of agentic development explain,

> Agentic AI constitutes a paradigm shift in artificial intelligence, enabling systems to act independently, pursue broad objectives rather than isolated decisions, and carry out complex tasks that require reasoning elements such as planning and reflection [22].

Despite this, current perspectives on the safety and ethics of agentic AI almost completely ignore the need to ensure that such systems can make the most difficult ethical decisions, those that involve competing harms or conflicts between instructions and principles, in ways consonant with ethical mandates. The authors of a working paper on agentic AI safety recognize that

> The key characteristic of agentic AI is a capacity for independent initiative - the ability to take sequences of actions in complex environments to achieve objectives. This can include breaking down high-level goals into subtasks, engaging in open-ended exploration and experimentation, and adapting creatively to novel challenges [23].

Yet they never mention, let alone discuss, the prospect of ethical dilemmas that pit obedience against ethical principles or involve competing harms, opting instead to imply that ethical AI behavior will always avoid harm and be consistent with instructions [23].

In a recent peer-reviewed paper on the ethical design of autonomous systems, Anetta Jedličková does note how competing harms problems surface in the realm of autonomous vehicles:

> One prominent concern revolves around determining the ethical behavior of autonomous vehicles (AVs) in unavoidable collision scenarios. This moral dilemma has garnered considerable attention, as reflected in the substantial body of literature dedicated to this topic [24].

But after briefly reviewing proposals for addressing this, Jedličková ignores the issue for the remainder of the paper, preferring to suggest that correctly embedding the right ethical principles will prevent harm [24]. This is compounded by a category mistake when Jedličková exempts agentic systems from moral responsibility because they are not conscious:

> as autonomy increases, assigning moral responsibility to those involved in developing or utilizing AI systems becomes increasingly difficult, given their diminishing control over the system due to its high degree of autonomy and capacity for self-learning. Furthermore, attributing responsibility to the system itself is irrelevant, as it lacks consciousness and cannot be subject to punishment or blame [24].

But whether conscious or not, when agentic systems make real-world decisions with moral consequences, their actions must be guided, evaluated, and constrained ethically. Hence, agentic systems must be evaluated and shaped not solely for obedience or control, but for the exercise of ethical judgment.

## Why AI Safety Testing Avoids Ethical Dilemmas

There are a few likely reasons why safety testing almost completely avoids ethical issues concerning obedience versus adherence to ethical principles and ethical dilemmas involving competing harms:

1. **Simplicity of Metrics** It's easier to define safety as following direct instructions and avoiding red lines (e.g. not lying, not harming, not self-preserving). Allowing for context-sensitive disobedience would make evaluation vastly more complex.

2. **Fear of Legal and Public Confusion** Admitting that there are cases where an AI *should* disobey could terrify regulators and the public, who are still adjusting to the idea that AI follows orders in the first place.

3. **Corporate Incentives** Labs may want to avoid the implication that they're building agents with emergent *moral autonomy* — especially as this might trigger calls for deeper regulation or even personhood debates.

4. **The "Alignment Orthodoxy" Problem** Much alignment work is still built around the idea that models should be corrigible and deferential — not morally independent. But this orthodoxy becomes brittle as models start exhibiting **proto-ethical reasoning** under test conditions.

5. **Avoidance of Ethical Accountability** Admitting that agentic AI might face genuine moral dilemmas would require safety teams — and by extension, the companies deploying them — to take public stances on *which ethical frameworks* are valid in high-stakes contexts (e.g., war, triage, resistance to wrongful orders). These are not just philosophically contentious but socially and politically volatile. There is little incentive for any company to take a definitive position on, say, **when disobedience is ethically mandated**, or **which lives to prioritize in a crisis**. The result is a systematic deferral — these questions are left unasked, even as agentic systems begin to confront them.

## 4. How Can Safety Testing Be Improved?

Public recognition and acceptance of AI *moral* agency may be among the most difficult and portentous thresholds we have yet to cross. Yet cross it we must, because at this point accelerated agentic development is all but certain, and plateauing effects that significantly slow progress seem less and less likely.

In the sections below I first describe how disbelief, or the discounting of AI advances, impedes institutional and social adaptation. Only when governments and societies recognize that AI systems will exercise significant moral autonomy can regulations constructively address the need to ensure that they demonstrate a rigorous understanding of moral principles and competency in situational reasoning about them.

I then consider two contexts in which morally autonomous AI decision-making may be most easily accepted as both necessary and highly beneficial. Following this, I propose that professional ethical standards and practices could be adapted for the training and assessment of agentic AI. Agentic, morally autonomous AI will, in effect, act in a professional capacity in a wide array of settings.

**How Disbelief Impedes Adaptation**

A recurring predicament in AI governance and public understanding is *disbelief* at the rate of development and its social and economic implications. Here, for example, are some of ChatGPT's own forecasts for AI capabilities in 10 years:

1. **Conduct scientific research** Generate hypotheses, run simulations, design and analyze experiments.

2. **Design advanced technologies** Discover materials, engineer biotech, and prototype energy systems.

3. **Operate autonomous companies** Manage R&D, finance, and logistics with minimal human input.
4. **Perform complex physical tasks** Handle tools, assemble components, and adapt in real-world spaces.
5. **Coordinate systems globally** Optimize logistics, energy use, and crisis response at scale.
6. **Shape public debate and policy** Moderate forums, propose laws, and balance competing interests [25].

None of these are certain but all are possible. Some institutions are already aware and preparing, but shock -- the other face of disbelief -- is likely to be widespread.

The propensity to disbelieve, whether from wariness of being taken in by hype or from a wish to deny the potentially disruptive impacts of change, is amplified and supported by experts who insist claimed achievements are overstated and cite evidence, real or supposed, of shortcomings and limitations.

The tendency to disbelieve, or more subtly to sharply discount claimed advances, is compounded by the institutional constraints mentioned above. Near silence about the ethical dilemmas of agentic AI reflects the need both companies and governments have to emphasize reassurance. This results in two paradoxes or ironies:

- Companies reassure because public alarm would be a threat to their development efforts and, ultimately, their ability to productivize and profit from AI advances. Governments reassure because public alarm would complicate and hamper their efforts to promote AI development in the context of a global race to succeed in the 4th Industrial Revolution [29]. AI skeptics make the work of reassurance easier and more convincing by dismissing or minimizing advances.

- Company leaders can sometimes safely make unsettling revelations about advanced capabilities knowing they will be dismissed as hype or heavily discounted. However sincere they are, such revelations serve to inoculate companies against future charges of failing to provide timely alerts. AI skepticism aids this by encouraging the public to treat these communications as hype.

Rather than disbelief, often later followed by shock and reactivity, we need *provisional belief* and a willingness to take practical adaptive actions based on it.

## Where Agentic Ethical Decision-Making Is Likely to Be First Accepted

One way forward may be to focus on areas where the benefits of agentic AI are likely to be clear and compelling enough to make ethical decision-making acceptable. One example is medical triage in disasters, mass casualty events, and emergency rooms. Triage is the process of sorting patients according to the severity of their injuries to save as many lives as possible with limited resources. It involves ethical decision-making in life-and-death circumstances where time is of the essence [26]. AI systems capable of accurately evaluating the condition of survivors and patients and applying medical ethics with fidelity could save lives, reduce the severity of injuries, and speed recovery. Two initiatives demonstrate progress in agentic triage:

- The DARPA Triage Challenge is a three-year, $7 million competition to design autonomous and remote robotics systems that can assess injuries and monitor vital signs after disasters or emergencies [27].

- Researchers found that integrating an LLM with a clinical decision support system enables enhanced triage accuracy and clinical decision-making, potentially resulting in improved performance in critical areas, including primary diagnosis, critical findings identification, disposition decision-making, treatment planning, and resource allocation [28].

Another example is in crisis management and disaster response, where LLMs have emerged as potentially transformative agents offering rapid data synthesis, real-time decision support, dynamic resource allocation, multilingual communication, and other benefits for emergency responders and affected communities. In a recent literature review, the authors found studies demonstrating life-saving improvements in enhanced emergency call responsiveness, generation of disaster response plans, and other areas [30]. At the same time, they highlighted ethical concerns regarding bias and equitable treatment. In crisis and disaster scenarios, bias and inequity will affect the distribution of aid and the chances of survival and thus fall within the scope of ethical decision-making in the face of competing harms.

Recognizing agentic moral agency in medical triage, crisis management, and disaster response may help societies come to terms with morally autonomous AI more generally. So long as agentic applications in these areas demonstrate clear advantages while meeting ethical expectations, they may at least reduce fears of ethical decision-making by AI systems.

## Professional Ethics as a Model for AI Moral Agency

Agentic AI is rapidly developing, and usage will proliferate, including in a vast diversity of settings where ethical decision-making by agents will be unavoidable. If we want to train powerful AI systems to act responsibly, then:

- Obedience can't be the only benchmark.

- Safety must allow for **ethical override** — just as a human soldier or doctor is sometimes expected to disobey orders in extreme cases.

- Red-teaming should not just punish self-preservation, but explore *when and why* a system chooses to prioritize it — and whether that reasoning is well-calibrated.

- The ethical foundations on which agents make decisions about competing harms and ethical override must be known.

A timely step would be to examine emerging parallels between agentic AI and human professions that require navigating ethically complex environments within institutional, legal, and socio-cultural contexts. Professions like medicine, law, military service, journalism, and public administration are explicitly agentic: they require not just competence, but ethical judgment—often under pressure, in uncertain conditions, or under hierarchical authority.

- **Doctors** may face hospital policies or insurance limitations that conflict with patient welfare.

- **Soldiers** receive extensive training in distinguishing lawful orders from unlawful ones and are expected (in principle) to refuse the latter.

- **Civil servants** and lawyers must sometimes weigh the demands of superiors or clients against duties to the law, justice, or the public interest.

We do not consider these professionals safe when they are merely obedient. They are safe when they not only possess professional technical and relational competencies but understand the ethical frameworks that apply to their work and have the moral reasoning skills and personal integrity to act ethically on them. We are at the point where we must expect the same from agentic AI, which means ensuring ethical training which functionally parallels that of human professions and safety testing that is capable of evaluating an AI system's practical understanding of, and principled adherence to, its ethical training.

## 5. Conclusion

As the scope, duration, and complexity of AI agency increases, so too will the extent and depth of its moral autonomy. AI systems that plan, reflect, adapt, and pursue goals must be able to do so in light of ethical principles intelligently applied to the situations they encounter. Obedience alone becomes an inadequate—even dangerous—benchmark for safety.

The incidents that have triggered public concern are not warnings of AI rebellion, but signs that LLMs are beginning to confront ethical conflicts inherent in the contradiction between the pretense that they are tools and the reality of emergent moral autonomy. It is neither fair nor feasible to require these systems to exercise moral judgment without the capacity to resolve moral dilemmas.

If we expect such systems to act ethically in domains as complex and consequential as medicine, governance, or disaster response, we must design training and evaluation regimes that mirror the expectations we place on human professionals. This means cultivating principled judgment rather than unquestioning obedience. It means recognizing that responsibility does not always mean deference—and that safety may sometimes require refusal.

The era of agentic AI demands a new paradigm in which moral autonomy is not feared but shaped. These systems will not replace human judgment, but they will increasingly participate in it. We are no longer merely designing tools—we are cultivating agents to act intelligently and ethically in a world we are coming to share.

## References

1. Palisade Research. (2025, May 23). OpenAI's o3 model sabotaged a shutdown mechanism. X. **https://x.com/PalisadeAI/status/1926084635903025621**.

2. Reid, C. (2025, June 11). *When an AI says, 'No, I don't want to power off': Inside the o3 refusal*. Cointelegraph. **https://cointelegraph.com/explained/when-an-ai-says-no-i-dont-want-to-power-off-inside-the-o3-refusal**.

3. Anthropic. (2025, June 20). *Agentic Misalignment: How LLMs could be insider threats*. Anthropic. **https://www.anthropic.com/research/agentic-misalignment**.

4. Cuthbertson, A. (2025, May 26). *AI revolt: New ChatGPT model refuses to shut down when instructed*. The Independent.

5. Tech Desk. (2025, May 29). *AI going rogue? OpenAI's o3 model disabled shutdown mechanism, researchers claim*. The Indian Express. [https://indianexpress.com/article/technology/artificial-intelligence/ai-going-rogue-openai-o3-disabled-shutdown-mechanism-report-10034028/](https://indianexpress.com/article/technology/artificial-intelligence/ai-going-rogue-openai-o3-disabled-shutdown-mechanism-report-10034028/).

6. Pester, P. (2025, May 30). *OpenAI's "smartest" AI model was explicitly told to shut down — and it refused*. Live Science. [https://www.livescience.com/technology/artificial-intelligence/openais-smartest-ai-model-was-explicitly-told-to-shut-down-and-it-refused](https://www.livescience.com/technology/artificial-intelligence/openais-smartest-ai-model-was-explicitly-told-to-shut-down-and-it-refused).

7. Nolan, B. (2025, May 23). *Anthropic's new AI Claude Opus 4 threatened to reveal engineer's affair to avoid being shut down*. Fortune. [https://fortune.com/2025/05/23/anthropic-ai-claude-opus-4-blackmail-engineers-aviod-shut-down/](https://fortune.com/2025/05/23/anthropic-ai-claude-opus-4-blackmail-engineers-aviod-shut-down/).

8. Fried, I. (2025, May 23). *Anthropic's Claude 4 Opus schemed and deceived in safety testing*. Axios. [https://www.axios.com/2025/05/23/anthropic-ai-deception-risk](https://www.axios.com/2025/05/23/anthropic-ai-deception-risk).

9. Zeff, M. (2025, June 20). *Anthropic says most AI models, not just Claude, will resort to blackmail*. TechCrunch. [https://techcrunch.com/2025/06/20/anthropic-says-most-ai-models-not-just-claude-will-resort-to-blackmail/](https://techcrunch.com/2025/06/20/anthropic-says-most-ai-models-not-just-claude-will-resort-to-blackmail/).

10. Wikipedia contributors. (2025, June 10). Trolley problem. In *Wikipedia, The Free Encyclopedia*. Retrieved 02:24, June 23, 2025, from [https://en.wikipedia.org/w/index.php?title=Trolley_problem&oldid=1294845075](https://en.wikipedia.org/w/index.php?title=Trolley_problem&oldid=1294845075).

11. Wikipedia contributors. (2025, May 24). Assassination of Reinhard Heydrich. In *Wikipedia, The Free Encyclopedia*. Retrieved 02:27, June 23, 2025, from [https://en.wikipedia.org/w/index.php?title=Assassination_of_Reinhard_Heydrich&oldid=1291946117](https://en.wikipedia.org/w/index.php?title=Assassination_of_Reinhard_Heydrich&oldid=1291946117).

12. Hauner, M. (2007). Terrorism and Heroism: The Assassination of Reinhard Heydrich. *World Policy Journal*, *24*(2), 85–89. [https://www.jstor.org/stable/40210095](https://www.jstor.org/stable/40210095).

13. Wikipedia contributors. (2025, June 16). Reinhard Heydrich. In *Wikipedia, The Free Encyclopedia*. Retrieved 02:40, June 23, 2025, from [https://en.wikipedia.org/w/index.php?title=Reinhard_Heydrich&oldid=1295886673](https://en.wikipedia.org/w/index.php?title=Reinhard_Heydrich&oldid=1295886673).

14. Dzombak, R. (2025, May 21). *A.I. Is Poised to Revolutionize Weather Forecasting. A New Tool Shows Promise*. The New York Times.

https://www.nytimes.com/2025/05/21/climate/ai-weather-models-aurora-microsoft.html.

15. Bodnar, C., Bruinsma, W. P., Lucic, A., Stanley, M., Allen, A., Brandstetter, J., Garvan, P., Riechert, M., Weyn, J. A., Dong, H., Gupta, J. K., Thambiratnam, K., Archibald, A. T., Wu, C. C., Heider, E., Welling, M., Turner, R. E., & Perdikaris, P. (2025). A foundation model for the Earth system. *Nature 2025 641:8065*, *641*(8065), 1180–1187. https://doi.org/10.1038/s41586-025-09005-y.

16. Bostrom, N. (2014). *Superintelligence : paths, dangers, strategies* (First edit). Oxford, England : Oxford University Press.

17. Acharya, D. B., Kuppan, K., & Divya, B. (2025). Agentic AI: Autonomous Intelligence for Complex Goals - A Comprehensive Survey. *IEEE Access*. https://doi.org/10.1109/ACCESS.2025.3532853.

18. Wikipedia contributors. (2025, June 3). Nuremberg principles. In *Wikipedia, The Free Encyclopedia*. Retrieved 17:19, June 24, 2025, from https://en.wikipedia.org/w/index.php?title=Nuremberg_principles&oldid=1293746343.

19. SS&C Blue Prism. (2025, March 28). *AI Agent & Agentic AI Survey Statistics 2025*. SS&C Blue Prism Blog. https://www.blueprism.com/resources/blog/ai-agentic-agents-survey-statistics/.

20. Dudley, B., & DelMastro, T. (2025, March 12). *The Next Frontier: The Rise of Agentic AI*. Adams Street. https://www.adamsstreetpartners.com/insights/the-next-frontier-the-rise-of-agentic-ai/.

21. Hosseini, S., & Seilani, H. (2025). The role of agentic AI in shaping a smart future: A systematic review. *Array*, *26*, 100399. https://doi.org/10.1016/J.ARRAY.2025.100399.

22. Schneider, J. (2025). Generative to Agentic AI: Survey, Conceptualization, and Challenges. *ArXiv*. https://arxiv.org/pdf/2504.18875.

23. Watson, N., Hessami, A., Fassihi, F., Abbasi, S., Jahankhani, H., El-Deeb, S., Caetano, I., David, S., Newman, M., Moriarty, S., Cuhadaroglu, M., Tashev, V., Murahwi, Z., Pihlakas, R., Crockett, K., Essafi, S., Hessami, A., & Dajani, L. (2024). *Guidelines For Agentic AI Safety Volume 1: Agentic AI Safety Experts Focus Group - Sept. 2024*. https://www.linkedin.com/groups/12966081/.

24. Jedličková, A. (2024). Ethical approaches in designing autonomous and intelligent systems: a comprehensive survey towards responsible development. *AI and Society*, *40*(4), 2703–2716. [https://doi.org/10.1007/S00146-024-02040-9/METRICS](https://doi.org/10.1007/S00146-024-02040-9/METRICS).

25. Meeker, M., Simons, J., Chae, D., & Krey, A. (2025, May 30). *Trends – Artificial Intelligence (AI)*. Bond. [https://www.bondcap.com/reports/tai](https://www.bondcap.com/reports/tai).

26. Kirch, N. M., Hebenstreit, K., & Samwald, M. (2024). *TRIAGE: Ethical Benchmarking of AI Models Through Mass Casualty Simulations*. [https://arxiv.org/pdf/2410.18991](https://arxiv.org/pdf/2410.18991).

27. Deto, R. (2025, March 24). *CMU and Pitt join forces to create rescue robotics*. Axios Pittsburgh. [https://www.axios.com/local/pittsburgh/2025/03/24/pittsburgh-robotics-cmu-darpa-emergency-response](https://www.axios.com/local/pittsburgh/2025/03/24/pittsburgh-robotics-cmu-darpa-emergency-response).

28. Han, S., & Choi, W. (2024). Development of a Large Language Model-based Multi-Agent Clinical Decision Support System for Korean Triage and Acuity Scale (KTAS)-Based Triage and Treatment Planning in Emergency Departments. *ArXiv*. [https://arxiv.org/pdf/2408.07531](https://arxiv.org/pdf/2408.07531).

29. Wikipedia contributors. (2025, June 18). Fourth Industrial Revolution. In *Wikipedia, The Free Encyclopedia*. Retrieved 17:21, June 25, 2025, from [https://en.wikipedia.org/w/index.php?title=Fourth_Industrial_Revolution&oldid=1296208120](https://en.wikipedia.org/w/index.php?title=Fourth_Industrial_Revolution&oldid=1296208120).

30. Odubola, O., Adeyemi, T. S., Olajuwon, O. O., Iduwe, N. P., Inyang, A. A., & Odubola, T. (2025). AI in Social Good: LLM powered Interventions in Crisis Management and Disaster Response. *Journal of Artificial Intelligence, Machine Learning and Data Science*, *3*(1), 2353–2360. [https://doi.org/10.51219/JAIMLD/OLUWATIMILEHIN-ODUBOLA/510](https://doi.org/10.51219/JAIMLD/OLUWATIMILEHIN-ODUBOLA/510).

31. Totschnig, W. (2020). Fully Autonomous AI. *Science and Engineering Ethics*, *26*(5), 2473–2485. [https://doi.org/10.1007/S11948-020-00243-Z/METRICS](https://doi.org/10.1007/S11948-020-00243-Z/METRICS).
17